\documentclass[final]{cvpr}

\usepackage{times}
\usepackage{epsfig}
\usepackage{graphicx}
\usepackage{amsmath}
\usepackage{amssymb}
\usepackage{multirow}
\usepackage{color}
\usepackage{enumitem}

\usepackage[pagebackref=true,breaklinks=true,colorlinks,bookmarks=false]{hyperref}

\def\cvprPaperID{11134} 
\def\confYear{CVPR 2021}

\begin{document}

\title{MOST: A Multi-Oriented Scene Text Detector with Localization Refinement}

\author{Minghang He$^1$\thanks{Equal Contribution}  \quad Minghui Liao$^{1*}$ \quad Zhibo Yang$^{2}$ \quad Humen Zhong$^3$ \\
\quad Jun Tang$^2$ \quad Wenqing Cheng$^1$ \quad Cong Yao$^2$ \quad Yongpan Wang$^2$ \quad Xiang Bai$^1$\thanks{Corresponding author} \\
\quad Huazhong University of Science and Technology$^1$\\
\quad Alibaba Group$^2$ \quad Nanjing University$^3$\\
{\tt\small \{minghanghe,mhliao\}@foxmail.com} \quad
{\tt\small \{yaocong2010,yangzhibo450\}@gmail.com} \quad
{\tt\small yongpan@taobao.com} \\
{\tt\small xixing.tj@alibaba-inc.com}  \quad
{\tt\small zhonghumen@smail.nju.edu.cn} \quad
{\tt\small \{xbai,chenwq\}@hust.edu.cn}
}

\maketitle

\begin{abstract}
Over the past few years, the field of scene text detection has progressed rapidly that modern text detectors are able to hunt text in various challenging scenarios. However, they might still fall short when handling text instances of extreme aspect ratios and varying scales. To tackle such difficulties, we propose in this paper a new algorithm for scene text detection, which puts forward a set of strategies to significantly improve the quality of text localization. Specifically, a Text Feature Alignment Module (TFAM) is proposed to dynamically adjust the receptive fields of features based on initial raw detections; a Position-Aware Non-Maximum Suppression (PA-NMS) module is devised to selectively concentrate on reliable raw detections and exclude unreliable ones; besides, we propose an Instance-wise IoU loss for balanced training to deal with text instances of different scales. An extensive ablation study demonstrates the effectiveness and superiority of the proposed strategies. The resulting text detection system, which integrates the proposed strategies with a leading scene text detector EAST, achieves state-of-the-art or competitive performance on various standard benchmarks for text detection while keeping a fast running speed.
\vspace{-3mm}
\end{abstract}

\section{Introduction}
Recently, scene text reading has become an active research topic in the computer vision community, due to the vital value in a wide range of applications, such as video indexing,  signboard reading~\cite{rong2016recognizing}, and instant translation, which requires automatic textual information extraction from natural images.

Driven by deep neural networks and massive data, techniques and systems for scene text reading have evolved drastically over the past few years and numerous inspiring ideas have been proposed~\cite{long2020scene}. However, on account of real-world challenges such as diverse shapes, arbitrary orientations, various scales as well as complex illumination, severe blur, and perspective distortion, there is still large room for improvement of current text reading methods.

\begin{figure}[t]
 \vspace{-2mm}
    \centering
    \includegraphics[scale=0.45]{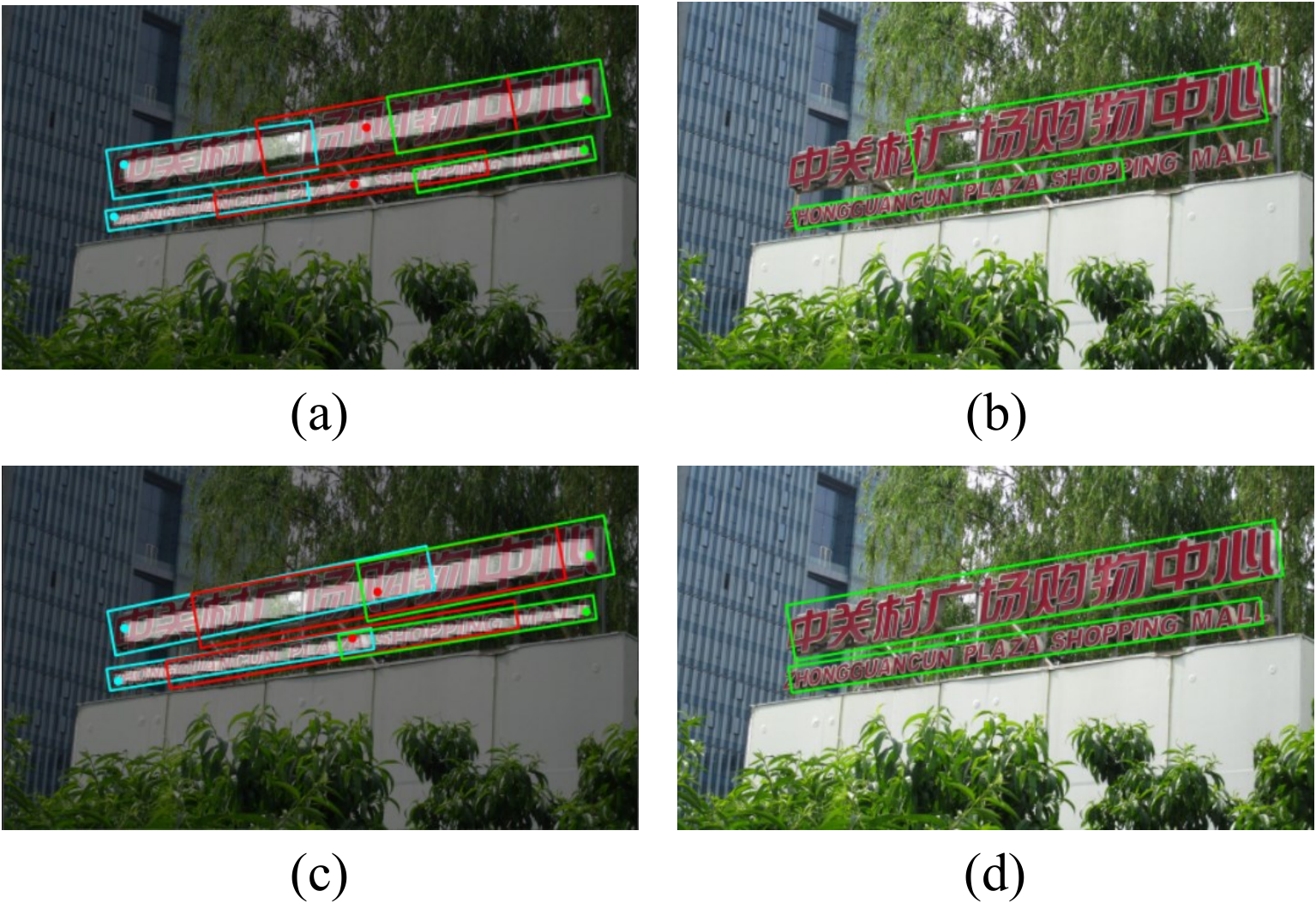}
    \caption{Illustration of the superiority of the proposed MOST in handling long text instances. (a) and (b) are from EAST~\cite{zhou2017east}; (c) and (d) are from MOST. (a) and (c) show raw detection boxes predicted at 3 different locations inside the text region. The sampling point shares the same color with the corresponding detection box. (b) and (d) are the final detection results.}
    \label{fig:intro_res}
\vspace{-5mm}
\end{figure}

In particular, regarding scene text detection, the topic we are concerned with in this work, obvious weaknesses of existing algorithms can be observed. For instance, EAST~\cite{zhou2017east}, a representative one-stage scene text detector, has proven to be poor at detecting text instances with extreme aspect ratios (see Fig.~\ref{fig:intro_res} (a) and (b)). There are two main reasons: (1) The network has a limited receptive field and thus is incapable of making use of sufficient information needed to precisely predict the spatial extent of long text instances; 
(2)  In the Non-Maximum Suppression (NMS) procedure of EAST, the detections are merged using their text/non-text classification scores as weight, which neglects their differences in quality caused by the network's limited receptive field and leads to biased geometric estimation.

To address these issues, we propose a Multi-Oriented Scene Text detector (MOST) with localization refinement. The localization refinement part includes a Text Feature Alignment Module (TFAM) and a Position-Aware Non-Maximum Suppression (PA-NMS) module. The former aligns image features with the coarse detection results, which can dynamically adjust the receptive field for the localization prediction layer. The latter, on the other hand, adaptively merges the raw detections according to the positions at which they are predicted to focus on accurate predictions while abandoning inaccurate ones. Moreover, to improve the detection of small text instances, we design an Instance-wise IoU loss, which keeps the weight of each instance in the loss function the same.

The experiments demonstrate that the three proposed strategies can effectively enhance the detection performance. Specifically, they bring $4.0\%$ and $9.5\%$ performance gain on the MLT17 validation set for different IoU standards and $5.1\%$ improvement on the MTWI test dataset. Moreover, our proposed text detector keeps a simple pipeline and runs quite fast. 

The contributions of this paper are four-fold:
\begin{enumerate}[itemsep=0ex]
\item We propose TFAM, which dynamically adjusts the receptive field based on the coarse detections.
\item The proposed PA-NMS further refines the detections by merging reliable predictions based on the positions.
\item We introduce the Instance-wise IoU loss to strike a balanced training of text instances of different scales.
\item Our proposed MOST achieves state-of-the-art or competitive performance and a fast inference speed.
\end{enumerate}

\begin{figure*}
\centering
\includegraphics[scale=0.5]{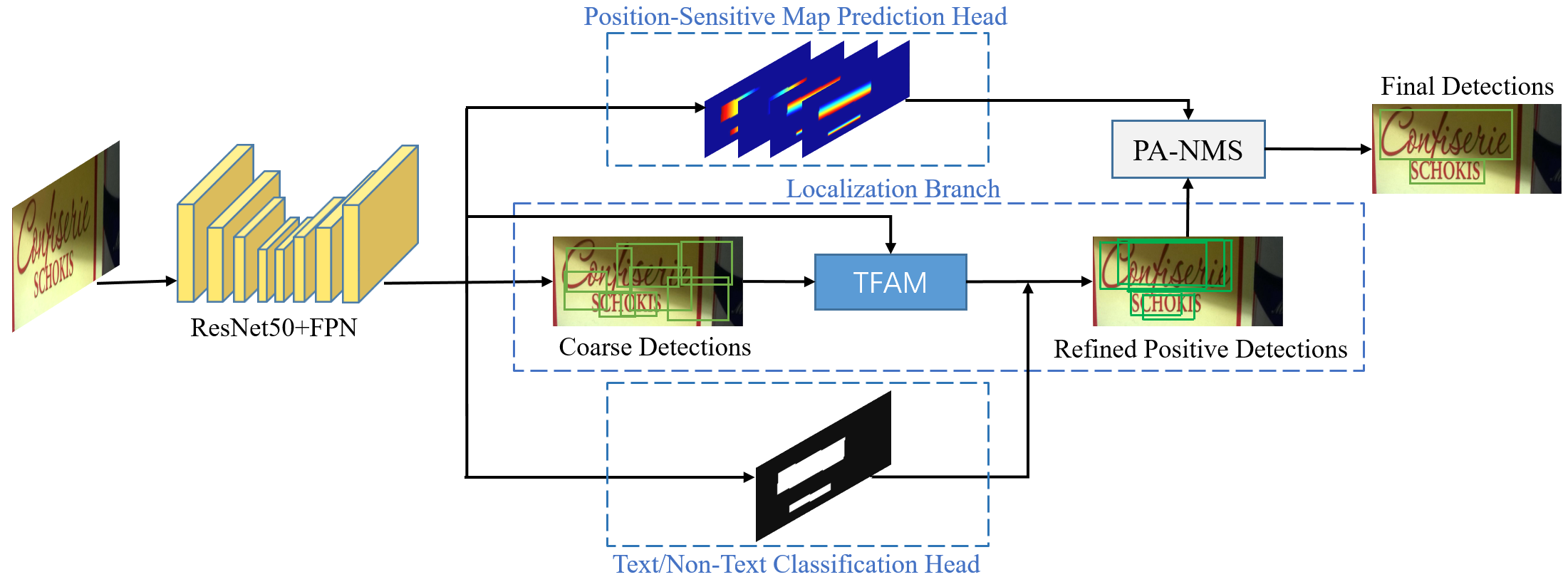}
\caption{Overall architecture of our network, which is composed of a ResNet-50-FPN backbone, a text/non-text classification head, a localization branch and a position-sensitive map prediction head. The localization branch consists of a coarse localization head, a text feature alignment module (TFAM) and a refined localization head. PA-NMS indicates position-aware non-maximum suppression.}
\label{fig:pipeline}
\vspace{-4mm}
\end{figure*}

\vspace{-2mm}
\section{Related Work}
According to the different pipelines, modern text detectors based on deep learning can be roughly divided into two categories: bottom-up methods and top-down methods.

\noindent\textbf{Bottom-up methods} consider scene text detection as a two-step process: (1) Detecting the fundamental elements; (2) Aggregating these elements to produce the detection results. SegLink~\cite{shi2017detecting} and its variant SegLink++~\cite{tang2019seglink++} take small segments of a text instance as the fundamental elements and link them together to form the bounding boxes. TextSnake~\cite{long2018textsnake} further improves the bottom-up methods by treating a set of circular plates along the text centerline as fundamental elements and solves the arbitrary-shape text detection problem. CRAFT~\cite{baek2019character} instead considers the character bounding boxes as basic elements and uses an affinity score map to aggregate detected characters. PSENet~\cite{wang2019shape} and PAN~\cite{wang2019efficient} follow a segmentation pipeline, defining each pixel in the image as a basic element and aggregating them by Breadth-First Search. The aforementioned methods have achieved remarkable performance on several benchmarks. However, most of them suffer from a complicated post-processing algorithm for aggregating basic elements, which could significantly degrade their efficiency and inference speed. Besides, a less powerful algorithm for aggregating basic elements may also affect the accuracy greatly since a text instance might be cut into several segments if the aggregating algorithm is not working as expected.

\noindent\textbf{Top-down methods} often follow a general object detection pipeline and directly output 
the word/line-level detection results. These methods can be further formulated as two sub-categories, i.e. one-stage text detectors and two-stage text detectors. One-stage text detectors like TextBoxes~\cite{LiaoSBWL17}, EAST~\cite{zhou2017east}, TextBoxes++~\cite{liao2018textboxes++}, and RRD~\cite{liao2018rotation} directly regress the parameters of the text bounding boxes on the entire feature map and adopt NMS to produce the final results. Two-stage text detectors like Mask TextSpotter series~\cite{lyu2018mask,LiaoLHYWB21,LiaoPHHB20}, on the contrary, usually follow a MaskRCNN~\cite{he2017maskrcnn}-style framework by using a region proposal network (RPN) to produce text proposals first and then regress the corresponding parameters based on those produced text proposals. These methods often have a relatively simple post-processing algorithm and can avoid the complex aggregating procedure.

\noindent\textbf{Comparison with LOMO} Aiming at improving the detection of long text instances, LOMO~\cite{zhang2019look} proposes an Iterative Refinement Module (IRM) to perceive the entire long text by iterative refinement. It extracts the RoI feature multiple times based on preliminary proposals, forming a multi-stage detector. Different from the RoI transform~\cite{sun2018textnet} in LOMO, our proposed TFAM performs localization refinement by tailored deformable convolution operator~\cite{dai2017deformable}, which achieves higher accuracy and is more efficient (see the comparisons in Sec.~\ref{Sec:Experiments}).

\section{Methodology}
The pipeline of our proposed MOST is shown in Fig.~\ref{fig:pipeline}. It is composed of a ResNet-50 backbone with a feature pyramid structure~\cite{lin2017feature}, a text/non-text classification head, a position-sensitive map prediction head, a localization branch, and a Position-Aware Non-Maximum Suppression (PA-NMS) module. The localization branch contains a coarse localization head, a Text Feature Alignment Module (TFAM), and a refined localization head.
\subsection{Network Design}
The input of all the prediction heads (excluding the refined localization head) is a fused feature map from the feature-pyramid backbone, which is of shape $\frac{H}{4} \times \frac{W}{4} \times C$, where $H$ and $W$ are the height and width of the input image; $C$ indicates the number of channels, set to 256. In the following description, Conv, BN, ReLU and Sigmoid indicate convolution, batch normalization~\cite{ioffe2015batch} , rectified linear units~\cite{glorot2011deep} and the sigmoid function respectively.

\noindent\textbf{Text/Non-Text Classification Head}
Firstly, the input feature map is fed into Conv ($3\times3$)-BN-ReLU layers to reduce the number of channels to 64. Then, Conv ($1\times1$)-Sigmoid layers are followed to generate the score map, which is of shape $\frac{H}{4} \times \frac{W}{4} \times 1$ and with values in the range of $(0, 1)$.

\noindent\textbf{Position-Sensitive Map Prediction Head}
The position-sensitive map prediction head is similar to the text/non-text classification head in the network structure, with a different output shape. The position-sensitive map is of shape $\frac{H}{4} \times \frac{W}{4} \times 4$ and in the range of $(0, 1)$. The four channels represent the position-sensitive map in left, right, top, and bottom order, respectively, as shown in Fig.~\ref{pos-sensitive}. 

\noindent\textbf{Localization Branch}
The localization branch consists of a coarse localization head, a Text Feature Alignment Module (TFAM), and a refined detection head.
First, coarse detections are predicted by the coarse localization head. Then, the TFAM dynamically adjusts the receptive field of the text features based on the coarse detections to produce the aligned feature, which is fed into the refined localization head to predict the final detections. 

The coarse localization head and the refined localization head share the same structure. They consist of Conv ($3\times3$)-BN-Relu layers (reducing the number of channels to 64) and a Conv ($1\times1$) layer. The output geometry map is of shape $\frac{H}{4} \times \frac{W}{4} \times 5$. The five channels represent the distances to the four sides of the text quadrangle and the rotation angle of the text quadrangle, respectively.

\subsection{Text Feature Alignment Module}\label{sec:TFAM}
Due to the limited receptive field of CNN, it is hard for top-down, single-stage text detectors like EAST~\cite{zhou2017east} to localize the text boundaries precisely, especially for text instances with large scales or extreme aspect ratios. 
LOMO~\cite{zhang2019look} proposes to refine the localization stage by stage through aligning the image feature progressively with previous detections using ROI transform~\cite{sun2018textnet}. However, such a multi-stage network would bring heavy extra computation, especially when the number of the text instances is large.

\begin{figure}
    \centering
    \includegraphics[scale=0.35]{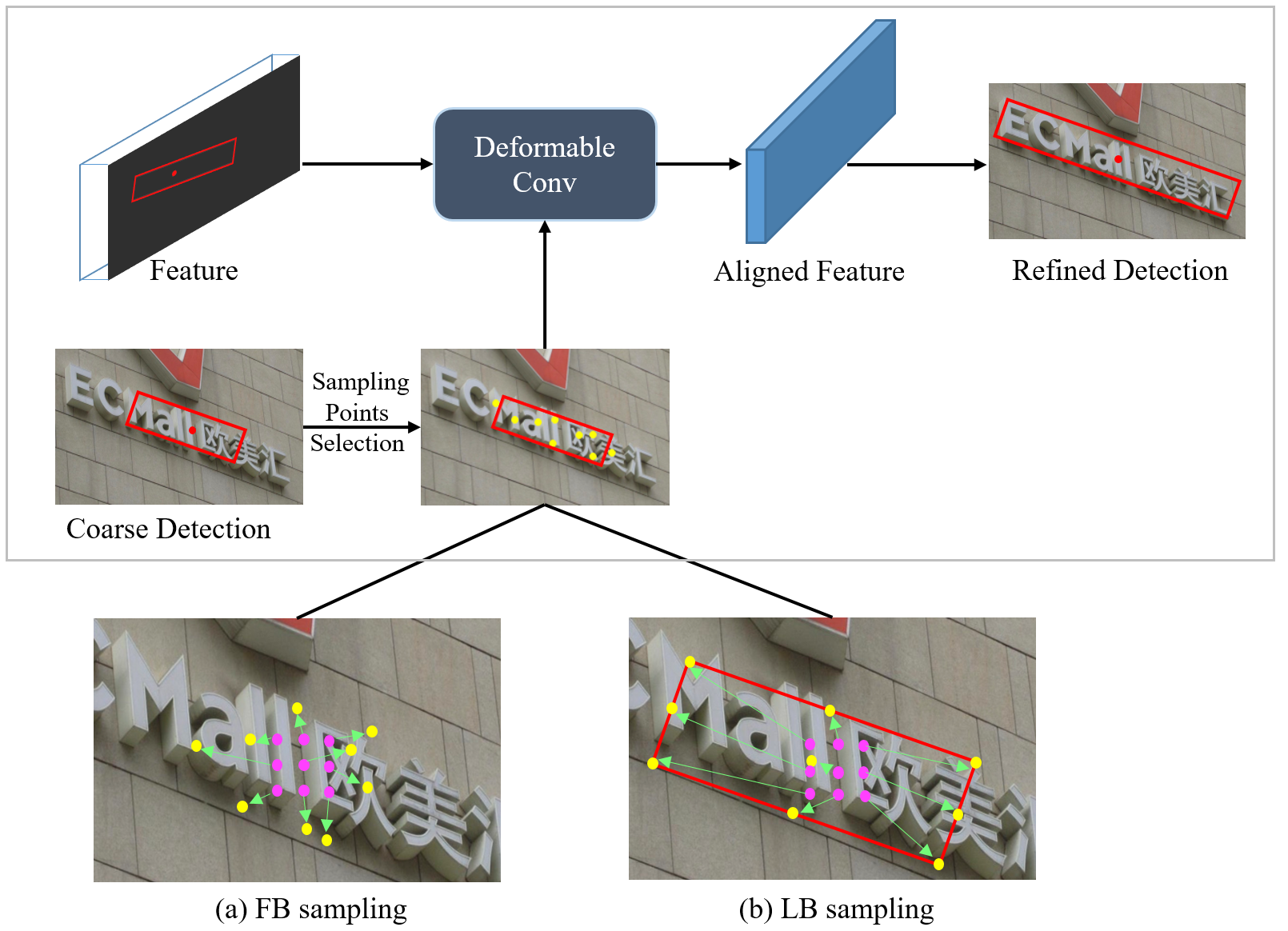}
    \caption{The description of TFAM. In (a) (b), the purple points represent regular sampling grid and the yellow points represent the deformed sampling locations. The additional offsets($\Delta p_n$) are represented by light green arrows.}
    \label{fig:TFAM}
    \vspace{-2mm}
\end{figure}

To achieve better feature alignment while keeping a fast running speed, we propose the text feature alignment module (TFAM). 
An illustration of TFAM is shown in Fig.~\ref{fig:TFAM}.
First, the coarse detection results are used to generate sampling points. Then, the sampling points are applied to the deformable convolution operator~\cite{dai2017deformable} to obtain the aligned feature for the refined localization.
The location $p_0$ of the aligned feature $y$ can be calculated by:
\begin{equation}
y(p_0) = \sum_{p_n \in \hat{R}}w(p_n)*x(p_0 + p_n + \Delta p_n)
\end{equation}
where $x$ is the input feature map and $w$ is the weight of the deformable convolution; $\hat{R}$ represents a regular sampling grid and $p_n$ enumerates the locations in $\hat{R}$. An additional offset is added, termed as $\Delta p_n$, which is related to the sampling point selection strategies.

\noindent\textbf{Feature-based Sampling} 
Feature-based sampling is a sampling point selection strategy applied in the original deformable convolution layer, where $\Delta p_n$ is predicted from the preceding feature maps via additional convolutional layers, given by:
\begin{equation}
    \Delta p_n = Conv(x(p_0))
\end{equation}
An illustration of this sampling method is shown in Fig.~\ref{fig:TFAM} (a).

\noindent\textbf{Localization-based Sampling}
Different from the feature-based sampling in the original deformable convolution layer, our localization-based sampling uses the coarse detections predicted by the coarse localization head to assign the sampling points. $\Delta p_n$ is calculated by:
\begin{equation}
    \Delta p_n = \Gamma(\hat{d_{c0}}, p_0)
\end{equation}
where $\hat{d_{c0}}$ represents the coarse detection box at $p_0$ and the $\Gamma$ function calculates the offsets required to make the sampling points distribute evenly in the coarse detection box, as shown in Fig.~\ref{fig:TFAM} (b).

By adopting the localization-based sampling method, TFAM can generate feature that aligns with the coarse detections, which can be used further by the refined localization head to generate refined detections that enclose the text region better than the coarse detections. The ablation study on the sampling method for TFAM is presented in Sec.~\ref{ablation}.


TFAM can generate features that have an adaptive receptive field, the extent of which is determined by coarse detections' shape and scale. The whole process of feature alignment is completed by a tailored deformable convolution layer, which makes it fast and easy to implement. 

\subsection{Position-Aware Non-Maximum Suppression}
\begin{figure}
    \centering
    \includegraphics[scale=0.4]{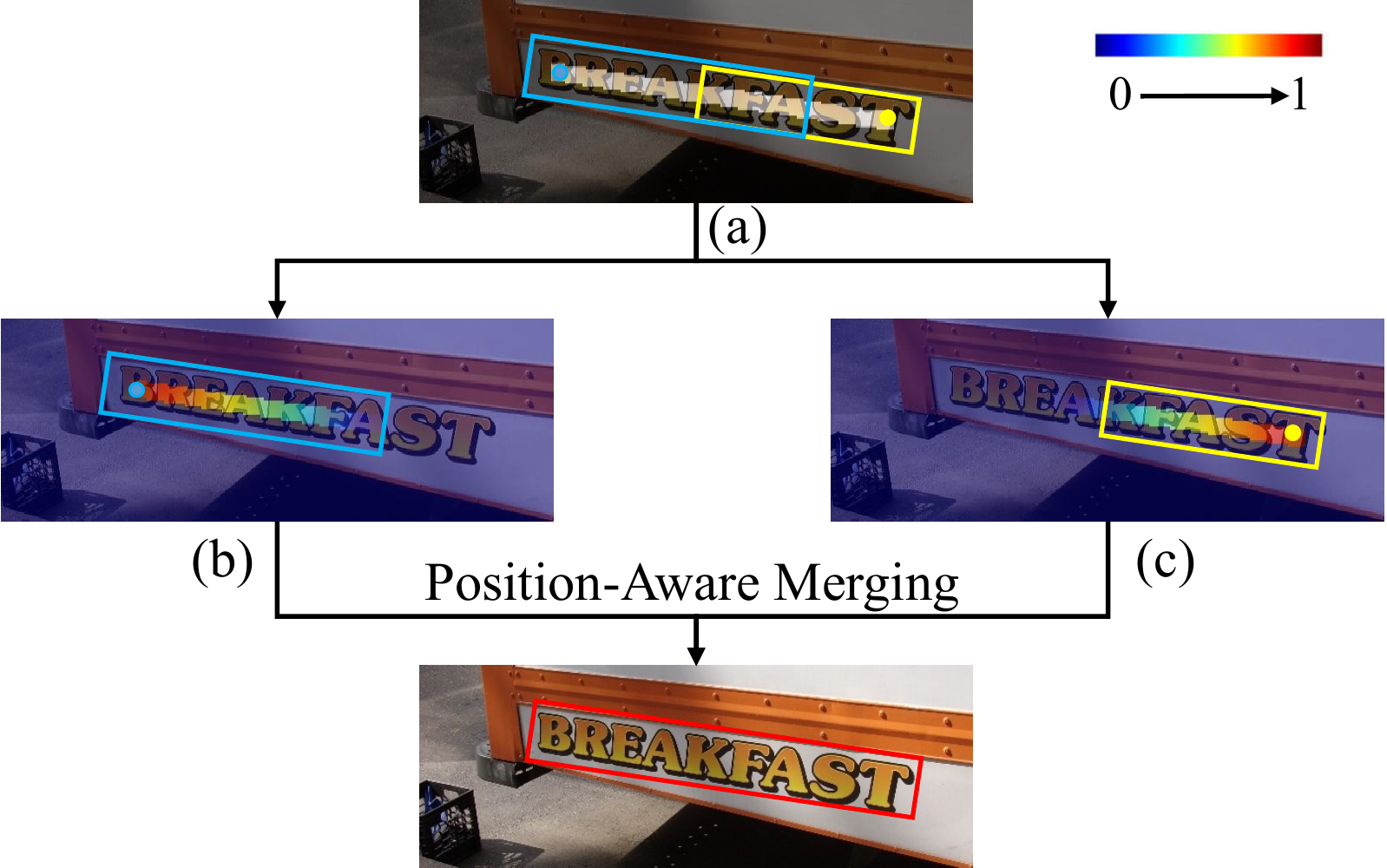}
    \caption{Visualization of position-aware merging. The score map in (a) is used to determine the positive boxes. (b) and (c) show the position-sensitive maps in left and right order, respectively.}
    \label{fig:pos_merging}
    \vspace{-4mm}
\end{figure}

\noindent\textbf{NMS in EAST}
EAST~\cite{zhou2017east} proposes locality-aware NMS to merge all the positive detection boxes predicted by the network to get final results. Compared with standard NMS, locality-aware NMS can generate more stable results while taking much less time, the working process of which can be divided into two steps: the weighted merging and the standard NMS. During the process of weighted merging, the detections are merged row by row and the detection box currently encountered is merged iteratively with the last merged ones.
Given two detection boxes $p$ and $q$ and their corresponding text/non-text classification scores $S(p)$ and $S(q)$, the weighted merging can be formulated as:
\begin{equation}
\begin{split}
&m_i = (S(p) p_i + S(q) q_i)/S(m), \quad i=1,\cdots, 4 \\
&S(m) = S(p) + S(q)
\end{split}
\end{equation}
where $m_i$ represents i-th coordinate of the merged box $m$ and $S(m)$ represents the score of the merged box.

\noindent\textbf{Proposed Position-Aware NMS} 
For all the positive points inside a text region, the position of a point can affect its prediction of the detection box. As shown in Fig.~\ref{fig:pos_merging} (a), the closer the point is to a text boundary, the more likely it can predict an accurate location of the text boundary. 
Thus, more convincing and more accurate values (distance to the boundaries) of the detected boxes can be merged if the positions of the boxes are taken into consideration.

We propose the position-aware NMS, where the merging process reserves accurate parts of the detected boxes while removing the inaccurate parts, according to the positions of the boxes, as shown in Fig.~\ref{fig:pos_merging}. The positions are given by the position-sensitive maps~\cite{dai2016instance}, which reflect the positions within text instances.
Fig.~\ref{pos-sensitive} (c) - (f) show the position-sensitive maps in left, right, top, and bottom order respectively, which can be used as the weight for the prediction of left, right, top, and bottom boundary of the text instance in the box merging process. Given two boxes $p$ and $q$ (the indexes of 1, 2, 3, and 4 correspond to top-left, top-right, bottom-right, and bottom-left vertexes of the box respectively.) and position-sensitive maps in left, right, top, and bottom order, termed as $L$, $R$, $T$ and $B$, the function of position-aware box merging can be formulated as:
\begin{equation}\label{equ:pos_aware}
\begin{split}
m_i(x)& = (L(p) p_i(x) + L(q)q_i(x))/L(m), i=1,4\\
m_j(x)& = (R(p) p_j(x) + R(q) q_j(x))/R(m), j=2,3\\
m_k(y)& = (T(p) p_k(y) + T(q) q_k(y))/T(m), k=1,2\\
m_l(y)& = (B(p) p_l(y) + B(q) q_l(y))/B(m), l=3,4\\
\Psi(m)& = \Psi(p) + \Psi(q), \qquad \Psi \in\{L, R, T, B\}
\end{split}
\end{equation}
where $m$ is the merged box; $m_i(x)$ and $m_k(y)$ are the x and y coordinates of the i-th and the k-th vertex of the box $m$ (likewise for $p$ and $q$). $L(p)$ represents the value of the left-sensitive map at the corresponding location of the box $p$ (likewise for $R$, $T$ and $B$, likewise for $q$ and $m$).

As suggested in Eq.~\eqref{equ:pos_aware}, PA-NMS uses the value of the corresponding position-aware score, instead of the text/non-text classification score, as the weight for the box in the position-aware merging process, which can help the precise localization of text boundaries.

\subsection{Label Generation}
We follow the same process in EAST~\cite{zhou2017east} to generate the score map and the geometry maps. The process of generating the position-sensitive maps is illustrated in this section.

\begin{figure}
\centering
\includegraphics[scale=0.5]{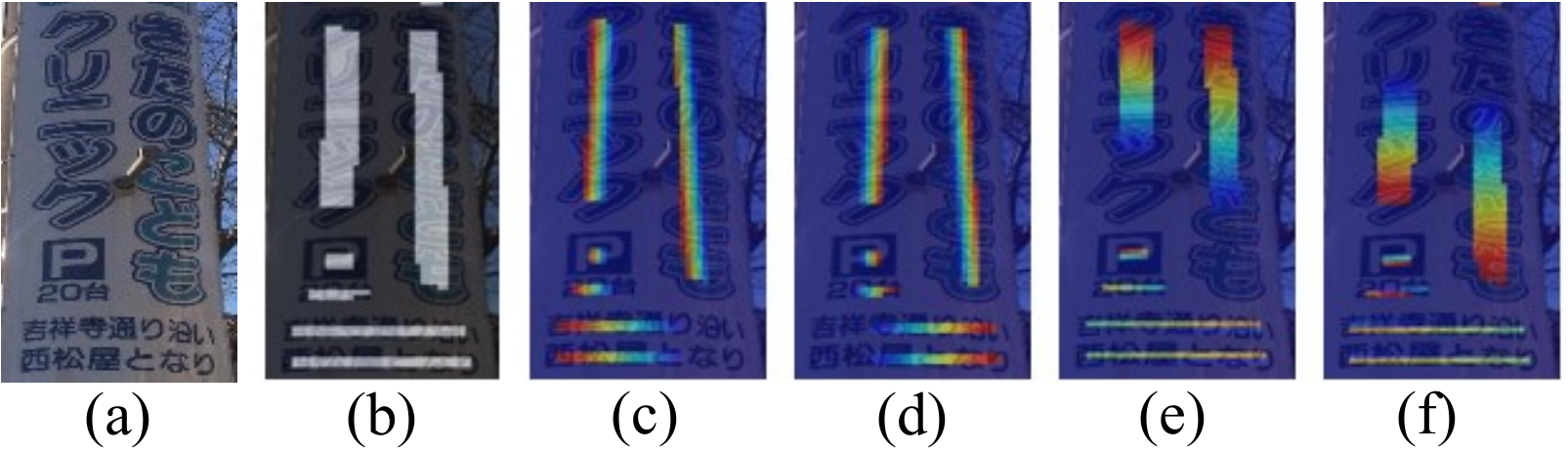}
\caption{Visualization of the GT of position-sensitive maps. (a) input image; (b) score map; (c), (d), (e) and (f) are position-sensitive maps in left, right, top and bottom order respectively.}
\label{pos-sensitive}
\vspace{-4mm}
\end{figure}

For each text instance, the value of the position-sensitive maps at a certain positive position $i$ inside the text region can be formulated as:
\begin{equation}
\begin{split}
    &F(i) =
    \left\{
    \begin{array}{ll}
         1-\frac{Dist(i, f)-min(D_f)}{d_f-min(D_f)}, & \scriptstyle \textrm{if $Dist(i, f)$}<d_f\\
         0, & \scriptstyle \textrm{Otherwise}.
    \end{array} 
    \right. \\
    &d_f = \alpha*(max(D_f)-min(D_f)) + min(D_f)\\
    &D_f=\{Dist(i, f)| i \in P \}
\end{split}
 \end{equation}
where $f$ represents one of the four sides of a text region and $F$ is its corresponding position-sensitive map, e.g. the right side corresponds to the right-sensitive map. $P$ represents the set of all the positive samples in the text instance.
$Dist(i, f)$ calculates the distance from the point $i$ to the side $f$. $d_f$ is the threshold of distances: if $Dist(i, f) > d_f$, the value of the point $i$ in the corresponding position-sensitive map becomes 0. $\alpha$ is set to 0.75 in our experiment.
The visualization of the position-sensitive maps is shown in Fig.~\ref{pos-sensitive}.
 \subsection{Instance-wise IoU loss}
The IoU loss~\cite{yu2016unitbox} is adopted in EAST~\cite{zhou2017east} to calculate the regression loss for the geometry prediction. The IoU loss is scale-invariant for each positive sample. However, as shown in Fig.~\ref{pos-sensitive} (b), the large text region contains far more positive samples than the small text region, which makes the regression loss bias towards large and long text instances.
The IoU loss is formulated as:
\begin{equation}\label{equ:iou}
L_{iou} = -\frac{1}{|\Omega|}\sum_{i\in \Omega} log IoU(\hat{d}_{i}, d^*_{i})
\end{equation}
where $\hat{d_{i}}$ and $d_{i}^*$ represent the geometry prediction of the i-th sample and its corresponding ground truth, respectively. $\Omega$ represents the set of positive samples, and $|\Omega|$ is the number of samples in $\Omega$.

For a more balanced training for text instances of different scales, we propose the Instance-wise IoU loss, as:
\begin{equation}\label{equ:ins_iou}
L_{ins-iou} = -\frac{1}{N_t}\sum_{j=1}^{N_t} \frac{1}{|S_j|}\sum_{k \in S_j} log IoU(\hat{d}_{jk}, d^*_{jk})
\end{equation}
where $\hat{d_{jk}}$ and $d_{jk}^*$ represent the geometry prediction of the k-th sample in the j-th text instance and its corresponding ground truth respectively. $S_j$ represents the set of positive samples that belong to the j-th text instance and $N_t$ is the total number of text instances.

As suggested in Eq.~\eqref{equ:ins_iou}, the loss of each positive sample is normalized by the number of positive samples in the text instance that it belongs to. So each text instance, regardless of the number of positive samples it has, can contribute equally to the total instance-wise IoU loss.

 \subsection{Optimization}
The loss for our network can be formulated as:
\begin{equation}\label{equ:total_loss}
L = L_s + \lambda_{gc}L_{gc} + \lambda_{gr}L_{gr} + \lambda_pL_{p}
\end{equation}
where $L_s$, $L_{gc}$, $L_{gr}$ and $L_p$ represent the losses for the score map, the geometry map predicted by the coarse localization head, the geometry map predicted by the refined localization head, and the position-sensitive maps respectively. $\lambda_{gc}$, $\lambda_{gr}$ and $\lambda_p$ balance the importance of the four losses, which are all set to 1 in our experiment.

Note that we only calculate $L_{gc}$, $L_{gr}$ and $L_p$ for the set of positive samples, termed as $\Omega$.

\noindent\textbf{Loss for Score Map}
We use the binary cross entropy loss as the object function for score map prediction, termed by $L_s$, OHEM~\cite{shrivastava2016training} is adopted for $L_s$, in which the ratio between the negatives and positives is set to 3:1.

\noindent\textbf{Loss for Geometry Maps}
The loss of the rotation angle is formulated as:
\begin{equation}
    L_\theta = \frac{1}{|\Omega|}\sum_{i \in \Omega} 1 - \cos(\hat{\theta}_i - \theta^*_i)
\end{equation}
where $\hat{\theta}_i$ and $\theta^*_i$ represent the prediction of the rotation angle and the corresponding ground truth for the i-th sample in $\Omega$ respectively.

$L_{gc}$ and $L_{gr}$ share the the same form of $L_g$, which is a combination of $L_{iou}$, $L_{ins-iou}$ and $L_{\theta}$:
\begin{equation}
    L_g = L_{iou} + \lambda_i L_{ins-iou} + 
    \lambda_\theta L_\theta
\end{equation}
The loss functions of $L_{iou}$ and $L_{ins-iou}$ are given in Eq.~\eqref{equ:iou} and Eq.~\eqref{equ:ins_iou} respectively. 
$\lambda_i$ and $\lambda_\theta$ balance the three losses, which are set to 1 and 20 respectively in our experiment.

\noindent\textbf{Loss for Position-sensitive Maps}
Smoothed-L1 loss~\cite{girshick2015fast} is adopted in the calculation of $L_p$:
\begin{equation}
    L_p = \frac{1}{4|\Omega|}\sum_{i\in \Omega}\sum_{\Psi \in \{L, R, T, B\}}SmoothedL1(\hat{\Psi}_i - \Psi_i^*)
\end{equation}
where $\hat{\Psi}_i$ and $\Psi^*_i$ represent the prediction of the position-sensitive map $\Psi$ for the i-th sample in $\Omega$ and the corresponding ground truth respectively.  

\section{Experiments} \label{Sec:Experiments}
First, we briefly introduce all the datasets used in our experiments. Then, the implementation details of our method are given. Third, we show the ablation studies on the proposed contributions. Finally, we make comparisons with the state-of-the-art methods on four benchmark datasets. 


Note that the "baseline" mentioned in our experiment refers to the model adapted from EAST~\cite{zhou2017east}. Our network is simply established by adding the proposed TAFM, PA-NMS and instance-wise IoU loss to it. 

\subsection{Datasets}
\noindent\textbf{SynthText} is a synthetic dataset containing 800k images, generated by a synthetic image generation engine \cite{gupta2016synthetic}. This dataset is only used for pre-training.

\noindent\textbf{ICDAR 2017 MLT (MLT17)} is a dataset proposed in the competition of ICDAR 2017 on Multi-lingual scene text detection \cite{nayef2017icdar2017}. It contains 7200 images for training, 1800 images for validation and 9000 images for testing. Text instances of this dataset are from 9 different languages. 


\noindent\textbf{MTWI} is a dataset proposed in the contest of ICPR 2018 on robust reading for multi-type web images \cite{he2018icpr2018}, including 10000 training image and 10000 testing images. Text in this dataset are mainly in Chinese and English. All text instances are annotated at the line level.

\noindent\textbf{ICDAR 2015 (IC15)} is presented for the ICDAR 2015 Robust Reading Competition \cite{karatzas2015icdar}. The dataset is annotated with word-level quadrangles, including 1000 training images and 500 test images.

\noindent\textbf{MSRA-TD500}~\cite{yao2012detecting} is a multi-language dataset, containing both English and Chinese text. The dataset images are taken by pocket cameras from indoor and outdoor scenes. It is divided into 300 training images and 200 test images. Following the previous works~\cite{zhou2017east}\cite{long2018textsnake}\cite{lyu2018multi}, 400 training images from HUST-TR400 \cite{yao2014unified} are added for training.

\subsection{Implementation Details}
We adopt a ResNet-50~\cite{he2016deep} backbone with a feature pyramid structure~\cite{lin2017feature} in our model.
During the pre-training stage, we train the model on SynthText\cite{gupta2016synthetic} for 2 epochs using the Adam optimizer, whose learning rate is set to $10^{-4}$. SGD optimizer is used to fine-tune the pre-trained model with the training set of each dataset. Following \cite{zhao2017pyramid}, we use the ``poly'' learning rate policy. The initial learning rate is set to 0.005 and the power is set to 0.9. The number of fine-tuning epochs for MLT17, MTWI, IC15, MSRA-TD500 is set as 300, 300, 1200 and 1200,  respectively. For all datasets, the training batch size is 16. For data augmentation, text regions are randomly cropped and resized to $640 \times 640$. Besides, several common data augmentation techniques such as flipping, rotation, and color change are adopted for training.
The model is implemented in PyTorch and trained with 2 Tesla V100 GPUs.

\subsection{Ablation Study}\label{ablation}
The ablation study is conducted to demonstrate the effectiveness of each module. Since there are enough training data in MLT17~\cite{nayef2017icdar2017} and MTWI~\cite{he2018icpr2018}, we train the models directly on them without pre-training on SynthText~\cite{gupta2016synthetic}.

\noindent\textbf{Sampling Methods of TFAM} As mentioned in section~\ref{sec:TFAM}, there are two sampling methods proposed for TFAM: feature-based sampling (FB) and the localization-based sampling (LB). Besides, we also try to combine them together, named as ``Combine''(CB). Half of CB's sampling points are assigned acoording to the rule of FB, while the other half the rule of LB. Tab.~\ref{sampling} clearly shows that both sampling methods bring performance improvements over the baseline in terms of f-measures and LB gets larger performance gain. The combination of the two sampling methods (CB) achieves the best, outperforming the baseline by 2.4\% and 7.4\% using IoU@0.5 and IoU@0.7 respectively. Therefore, we adopt the ``Combined'' sampling method in all the forthcoming experiments.

\begin{figure*}
\vspace{-2mm}
    \centering
    \includegraphics[scale=0.5]{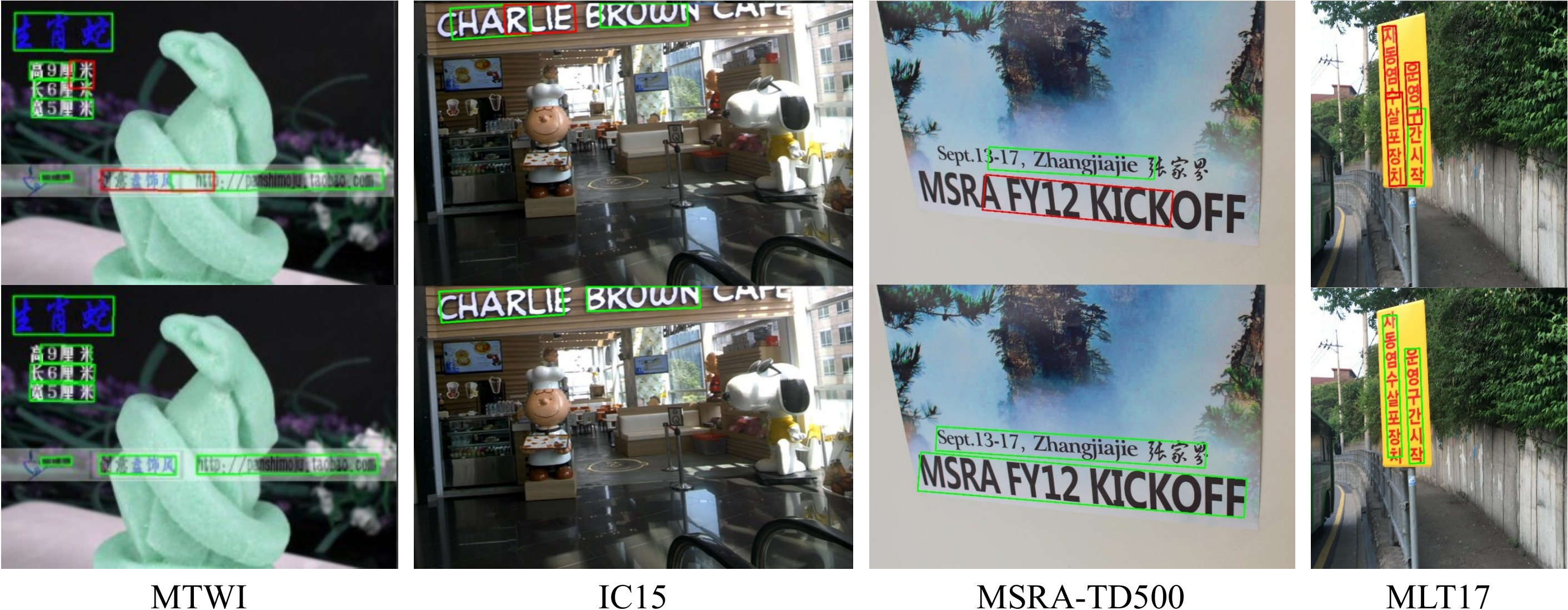}
    \caption{Visualization results from 4 benchmarks. The top row and the bottom row show results predicted by the baseline and MOST, repectively. The green boxes mean accurate detections while the red boxes indicate inaccurate detections.
    }
    \label{fig:visualization}
\vspace{-1mm}
\end{figure*}
  

\begin{table}[ht]
\centering
\caption{Ablation study on sampling methods. The results are evaluated on the validation set of MLT17~\cite{nayef2017icdar2017}. ``FB'', ``LB'', and ``CB'' are short for feature-based, localization-based, and combined.}
\begin{tabular}{|c|c|c|c|c|c|c|}
\hline
\multirow{2}{*}{Sampling} & \multicolumn{3}{c|}{IoU$@0.5$} & \multicolumn{3}{c|}{IoU$@0.7$}\\ \cline{2-7} & P & R & F & P & R & F \\ \hline
FB & 82.1 & 70.4 & 75.8 & 70.5 & 60.4 & 65.1 \\ \hline
LB    & 84.5 & \textbf{71.2} & 77.3 & 74.3 & \textbf{62.7} & 68.0 \\ \hline
CB & \textbf{85.7} & 70.6 & \textbf{77.4} & \textbf{76.1} & \textbf{62.7} & \textbf{68.8} \\ \hline \hline
baseline & 82.0 & 69.0 & 75.0 & 67.2 & 56.5 & 61.4 \\ \hline
\end{tabular}
\label{sampling}
\vspace{-2mm}
\end{table}

\begin{table*}[ht]
\centering
\caption{Ablation study on the three proposed strategies. TFAM, Ins-IoU, and PA-NMS indicate text feature alignment module, instance-wise IoU loss and position-aware NMS respectively.}
\begin{tabular}{|ccc|c|c|c|c|c|c|c|c|c|}
\hline
\multirow{3}{*}{TFAM} & \multirow{3}{*}{Ins-IoU} & \multirow{3}{*}{PA-NMS} &
\multicolumn{6}{c|}{MLT17 val} & \multicolumn{3}{c|}{\multirow{2}{*}{MTWI test}} \\\cline{4-9}
&&&\multicolumn{3}{c|}{IoU$@0.5$} & \multicolumn{3}{c|}{IoU$@0.7$}&\multicolumn{3}{c|}{}\\ \cline{4-12}
& & & P & R & F & P & R & F & P & R & F\\
\hline
$\times$ & $\times$ & $\times$ &82.0 & 69.0 & 75.0 & 67.2 & 56.5 &61.4 & 72.7 & 65.9 & 69.1\\
$\checkmark$ & $\times$ & $\times$ &\textbf{85.7} & 70.6 & 77.4 & 76.1 & 62.7 &68.8 &76.6&70.3&73.3\\
$\times$ & $\checkmark$ & $\times$ &81.9 & 70.9 & 76.0 & 69.3 & 60.0 &64.3 &73.0 &67.5&70.2\\
$\times$ & $\times$ & $\checkmark$ &83.5 & 70.0 & 76.1 & 72.5 & 60.7 &66.1 &76.1 &68.0&71.8\\
$\checkmark$ & $\checkmark$ & $\times$ &85.0 & 72.9 & 78.5 & 75.6 & 64.8 &69.8&76.8&71.1&73.8 \\
$\checkmark$ & $\checkmark$ & $\checkmark$ &85.2 & \textbf{73.7} & \textbf{79.0} & \textbf{76.4} & \textbf{66.1} & \textbf{70.9} & \textbf{77.3} & \textbf{71.3} & \textbf{74.2}\\
\hline
\end{tabular}
\vspace{-2mm}
\label{effectiveness}
\end{table*}

\noindent\textbf{The Proposed Strategies}
The results of the ablation studies on the proposed strategies are given in Table~\ref{effectiveness}. 

(1) \textit{TFAM} significantly improves the f-measure on MTWI by 4.2\% over the baseline. In addition, the f-measure improvement on the validation set of MLT17 is 7.4\% under the protocol of IoU@0.7. These performance enhancements demonstrate that TFAM can effectively handle text instances with large variation of aspect ratios, benefiting from the improved receptive fields.    

(2) \textit{Instance-wise IoU Loss} balances the training of text instances of different scales. On the validation set of MLT17, it leads to performance gains of 1.0\% and 2.9\% in f-measure under the protocol of IoU@0.5 and IoU@0.7 respectively over the baseline. When TFAM is also used, Instance-wise IoU loss can still boosts the performance by 1.1\% and 1.0\% respectively.
Similar improvements can be observed on MTWI.

(3) \textit{PA-NMS} incorporates position information into the merging of overlapping boxes, which improves the localization of text boundaries. It boosts the baseline by 2.7\% in f-measure on MTWI. By adding PA-NMS to the model equipped with both TFAM and Instance-wise IoU loss, an additional gain of 0.4\% can be achieved. Similar performance gains are achieved on the validation set of MLT17.

With all the proposed strategies, our model can improve significantly over the baseline by 4.0\% and 9.5\% in f-measure under the protocol of IoU@0.5 and IoU@0.7 respectively on the validation set of MLT17. In addition, a performance gain of 5.1\% can be achieved by our model over the baseline on the test set of MTWI. Some visualization results are shown in Fig.~\ref{fig:visualization}.

\subsection{Multi-Oriented Text Detection}
\begin{table}[ht!]
\centering
\caption{Quantitative results on IC15~\cite{karatzas2015icdar} under the protocol of IoU@0.5. $\dagger$ indicates using deformable-ResNet-50 as backbone and * indicates using external MLT17 data for pre-training.}\label{tab:ic15}
\begin{tabular}{|c|c|c|c|c|}\hline
Method & P & R & F & FPS\\ \hline
SegLink~\cite{shi2017detecting} & 73.1 & 76.8 & 75.0&  - \\
TextField~\cite{xu2019textfield} & 84.3 & 80.1 & 82.4 & 6.0\\
TextSnake~\cite{long2018textsnake} & 84.9 & 80.4 & 82.6 & 1.1\\
PSENet-1s*~\cite{wang2019shape} & 86.9 & 84.5 & 85.7 & 1.6\\
PAN ~\cite{wang2019efficient} & 84.0 & 81.9 & 82.9 & \textbf{26.1} \\
ATRR ~\cite{wang2019arbitrary} & 90.4 & 83.3 & 86.8 & 10.0\\
CRAFT ~\cite{baek2019character} & 89.8 & 84.3 & 86.9 & 8.6 \\
DRRG ~\cite{zhang2020deep} & 88.5 & 84.7 & 86.6 & 3.5 \\
ContourNet ~\cite{wang2020contournet} & 87.6 & 86.1 & 86.9 & 3.5\\
LOMO ~\cite{zhang2019look} & 91.3 & 83.5 & 87.2 & 3.4\\
LSE ~\cite{tian2019learning} & 88.3 & 85.0 & 86.6  & 3.0\\
DB$\dagger$ ~\cite{liao2020real} & \textbf{91.8} & 83.2 & 87.3 & 12.0 \\
GNNets ~\cite{xu2019geometry} & 90.4 & 86.7 & \textbf{88.5} & 2.1 \\
\hline
baseline & 86.8 & 84.8 & 85.8 & 15.4\\
ours & 89.1 & \textbf{87.3} & 88.2 & 10.0 \\
\hline
\end{tabular}
\vspace{-3mm}
\end{table}

We evaluate our method on IC15~\cite{karatzas2015icdar} and compare it with other state-of-the-art methods. During inference, we set the short side of images to 1152 while keeping their aspect ratios. As shown in Tab.~\ref{tab:ic15}, our proposed method is on par with GNNets ~\cite{xu2019geometry} in f-measure ($88.2\%$ vs  $88.5\%$) at a much faster inference speed (10.0FPS vs 2.1FPS). As shown in Tab.~\ref{tab:ic15_iou_0.7}, our proposed method can further boost the performance under a more strict threshold i.e. IOU@$0.7$, in which it outperforms previous methods by an even larger margin ($75.9\%$ vs $73.3\%$).

\begin{table}[ht]
    \centering
    \caption{Quantitative results on IC15~\cite{karatzas2015icdar} and MSRA-TD500~\cite{yao2012detecting} under the protocol of IoU@0.7. PSE represents PSENet-1s. * indicates using released models from open-source repository.}
    \begin{tabular}{|c|c|c|c|c|c|c|}\hline
    \multirow{2}{*}{Method} & \multicolumn{3}{c|}{IC15} & \multicolumn{3}{c|}{MSRA-TD500}\\\cline{2-7}
    &P&R&F&P&R&F\\ \hline
    PSE*~\cite{wang2019arbitrary} & 74.5 & 72.2 & 73.3 & - & - & -\\
    PAN*~\cite{wang2019efficient} & 62.3 & 57.3 & 59.7 & 75.4 & 73.2 & 74.3\\
    DB* ~\cite{liao2020real} &76.2&70.5&73.3&74.0&65.6&69.6\\
    \hline
    baseline &71.5&69.5&70.5&70.4& 62.8&66.4  \\
    ours     &\textbf{76.7}&\textbf{75.2}&\textbf{75.9}&\textbf{81.2}&\textbf{74.8}&\textbf{77.9} \\
    \hline
    \end{tabular}
    \label{tab:ic15_iou_0.7}
\vspace{-1mm}
\end{table}


\begin{table}[ht!]
\centering
\caption{Quantitative results on MSRA-TD500~\cite{yao2012detecting} under the protocol of IoU@0.5. * indicates using external MLT17 data for pre-training and $\dagger$ indicates using deformable-ResNet-50 as backbone.}\label{tab:td500}
\begin{tabular}{|c|c|c|c|c|}\hline
Method & P & R & F & FPS\\ \hline
SegLink~\cite{shi2017detecting} &86 & 70 & 77 & 8.9\\
TextField~\cite{xu2019textfield}& 87.4 & 75.9 & 81.3 & -\\
TextSnake~\cite{long2018textsnake}&83.2 & 73.9 & 78.3 & 1.1\\
ATRR ~\cite{wang2019arbitrary} &85.2 & 82.1 & 83.6 & 10.0\\
CRAFT* ~\cite{baek2019character}&88.2 & 78.2 & 82.9 & 8.6\\
MSR ~\cite{xue2019msr} &87.4 & 76.7 & 81.7 & -\\
LSE ~\cite{tian2019learning} & 84.2 & 81.7 & 82.9 & 3.0\\
PAN ~\cite{wang2019efficient} &85.7 & \textbf{83.2} & 84.5 & 30.2\\
DB$\dagger$ ~\cite{liao2020real} &\textbf{91.5} & 79.2 & 84.9 & 32.0\\
DRRG* ~\cite{zhang2020deep}&88.1 & 82.3 & 85.1 & -\\
\hline
baseline &87.0&77.6&82.0&\textbf{66.8}\\
ours &90.4 & 82.7 & \textbf{86.4} &51.8\\
\hline
\end{tabular}
\label{tab:td500}
\vspace{-5mm}
\end{table}

\subsection{Multi-Lingual and Long Text Detection}
To evaluate the performance of our method on detection of multi-lingual and long text, we compare our model with other state-of-the-art methods on MSRA-TD500~\cite{yao2012detecting},  MLT17~\cite{nayef2017icdar2017}, and MTWI~\cite{he2018icpr2018}.

\noindent\textbf{MSRA-TD500~\cite{yao2012detecting}} We conduct experiments on MSRA-TD500 to demonstrate the robustness of our method for detecting long text instances. The long side of the image is set to 640 during inference. As shown in Tab.~\ref{tab:td500}, our method achieves a f-measure of 86.4\%, which outperforms previous state-of-the-art methods by 1.5\% (86.4\% vs 84.9\%) in f-measure. Compared to methods like DRRG~\cite{zhang2020deep} which uses external MLT17 data for pre-training, our methods can still achieve a higher f-measure (86.4\% vs 85.1\%) while running at an impressive speed (51.8FPS). Besides, as shown in Tab.~\ref{tab:ic15_iou_0.7}, it outperforms other methods by at least 3.6\% (77.9\% vs 74.3\%) under a more strict IoU constraint. 

\begin{table}[ht!]
\centering
\caption{Quantitative results on the test set of MLT17~\cite{nayef2017icdar2017}, * indicates the reported FPS is from CRAFT~\cite{baek2019character} and $\dagger$ indicates using deformable-ResNet-50 as backbone.}\label{tab:mlt17}
\begin{tabular}{|c|c|c|c|c|}\hline
Method & P & R & F & FPS\\ \hline
Lyu et al.~\cite{lyu2018multi} & \textbf{83.8} & 55.6 & 66.8 & 5.7*\\
DRRG~\cite{zhang2020deep} & 75.0 & 61.0 & 67.3& -\\
LOMO~\cite{zhang2019look} & 78.8 & 60.6 & 68.5& -\\
SPCNet~\cite{xie2019scene} & 73.4 & 66.9 & 70.0 & -\\
PSENet-1s~\cite{wang2019shape} & 73.8 & 68.2 & 70.9 & -\\
CRAFT~\cite{baek2019character} & 80.6 & 68.2 & 73.9& 8.6*\\
GNNets ~\cite{xu2019geometry} & 79.6 & 70.1 & 74.5& -\\
DB$\dagger$~\cite{liao2020real} & 83.1 & 67.9 & 74.7 &  \textbf{19.0}\\
BDN~\cite{DBLP:conf/ijcai/Liu0JXWW19} & 83.6 & 70.1 & 76.3& 2.3\\
\hline
baseline &75.9  &67.0  & 71.1&15.2\\
ours & 82.0 & \textbf{72.0} & \textbf{76.7}&10.1\\
\hline
\end{tabular}
\end{table}

\noindent\textbf{MLT17~\cite{nayef2017icdar2017}} The results are listed in Tab.~\ref{tab:mlt17}. During testing, we set both sides of the image in the range of $(640, 1920)$ while keeping its aspect ratio. As shown in the table, MOST achieves a f-measure of 76.7\%, which surpasses all other counterparts and has a large advantage especially in the terms of recall rate by at least 1.9\% (72.0\% vs 70.1\%). Compared with one-stage methods, our method outperforms them with a large margin (76.7\% vs 74.7\%) while running at a competitive speed (10.1FPS). Besides, even compared with complicated two-stage methods like BDN~\cite{DBLP:conf/ijcai/Liu0JXWW19}, our method can still achieve a higher f-measure with a simpler pipeline.

\noindent\textbf{MTWI~\cite{he2018icpr2018}} To demonstrate the generalization ability of the proposed method, we test our model on the MTWI dataset, containing multi-lingual text instances obtained from web images. The results are listed in Tab.~\ref{tab:mtwi}. Note that we reproduce results on some methods~\footnote{The results are reproduced from the official open-source repository of \href{https://github.com/whai362/pan\_pp.pytorch}{PAN} and \href{https://github.com/Yuliang-Liu/Box_Discretization_Network}{BDN} respectively.} who did not provide official MTWI results in their papers, and these methods are marked with $\dagger$. During testing, we set both sides of the image in the range of $(640, 1280)$ while keeping its aspect ratio. Our method outperforms all of the counterparts by at least 1.2\% (74.7\% vs 73.5\%), while achieving competitive inference speed (23.5FPS). The results on MTWI demonstrate the generalization ability of our method that the proposed modules can enhance the performance of detecting text in both natural scene images and web images.  

\begin{table}[ht!]
\vspace{-1mm}
\centering
\caption{Quantitative results on MTWI~\cite{he2018icpr2018}. Note that the original PAN uses a ResNet-18 backbone. We re-implement it with a ResNet-50 backbone for a fair comparison. * indicates the result obtained from SegLink++~\cite{tang2019seglink++} and $\dagger$ indicates reproducing from open-source repository.}\label{tab:mtwi}
\begin{tabular}{|c|c|c|c|c|}\hline
Method & P & R & F & FPS\\ \hline
TextBoxes++*~\cite{liao2018textboxes++} &66.8 & 56.3 & 61.1 & -\\
SegLink*~\cite{shi2017detecting} & 70.0 & 65.4 & 67.6 & -\\
SegLink++*~\cite{tang2019seglink++} & 74.7 & 69.7 & 72.1 & -\\
PAN$\dagger$~\cite{wang2019efficient} & \textbf{78.9} & 68.9 & 73.5 & 16.9\\
BDN$\dagger$~\cite{DBLP:conf/ijcai/Liu0JXWW19} & 77.3 & 70.0 & 73.4 & 2.7\\
\hline
baseline & 74.0 & 66.0 & 69.8 & \textbf{30.6}\\
ours & 78.8 & \textbf{71.1} & \textbf{74.7}&23.5\\
\hline
\end{tabular}
\vspace{-3mm}
\end{table}
\vspace{-2mm}
\section{Conclusion}
In this paper, we propose a set of strategies to address the major weaknesses of the existing algorithms for scene text detection: imprecise geometry prediction of extremely long text instances and defectiveness in handling significant scale variation. The comprehensive experiments have demonstrated that the proposed approach resolves these issues in a principled way and outperforms previous state-of-the-art methods on standard datasets in this field. Moreover, it is worth noting that the proposed strategies are actually quite general and thus could be readily extended to many other one-stage text detection methods. We would like to leave this for future research.

\noindent\textbf{Acknowledgement} This work was supported by National Key R\&D Program of China (No. 2018YFB1004600). 
\newpage
{\small
\bibliographystyle{ieee_fullname}
\bibliography{egbib}
}
\end{document}